\documentclass{article}

\PassOptionsToPackage{square,numbers}{natbib}

\usepackage[preprint]{neurips_2025}

\usepackage[utf8]{inputenc} 
\usepackage[T1]{fontenc}    
\usepackage[hidelinks]{hyperref}       
\usepackage{graphicx}
\usepackage{url}            
\usepackage{booktabs}       
\usepackage{amsfonts}       
\usepackage{nicefrac}       
\usepackage{microtype}      
\usepackage[table]{xcolor}         
\usepackage{booktabs}
\usepackage{algorithm}
\usepackage{algorithmic}
\usepackage{tikz}
\usepackage{array}
\usepackage{makecell}
\usepackage{enumitem}
\usepackage{soul}
\usepackage{mathtools}
\usepackage{amssymb}
\usepackage{pgfplots}
\pgfplotsset{compat=1.17}
\usepgfplotslibrary{external,statistics}
\usepackage{wrapfig}
\usepackage{subcaption}
\usepackage{mdframed} 

\newmdenv[
  linewidth=.4pt,
  topline=true,bottomline=false,  
  leftline=true,rightline=true,
  innerleftmargin=6pt,innerrightmargin=6pt,
  innertopmargin=6pt,innerbottommargin=0pt
]{topframe}

\newmdenv[
  linewidth=.4pt,
  topline=false,bottomline=true,  
  leftline=true,rightline=true,
  innerleftmargin=6pt,innerrightmargin=6pt,
  innertopmargin=0pt,innerbottommargin=6pt
]{botframe}

\title{Using Reasoning Models to Generate Search Heuristics that Solve Open Instances of Combinatorial Design Problems}

\author{
    Christopher D. Rosin \\
    https://constructive.codes \\
    christopher.rosin@gmail.com
}

\begin{document}

\maketitle

\begin{abstract}

Large Language Models (LLMs) with reasoning are trained to iteratively generate and refine their answers before finalizing them, which can help with applications to mathematics and code generation.  We apply code generation with reasoning LLMs to a specific task in the mathematical field of combinatorial design.  This field studies diverse types of combinatorial designs, many of which have lists of open instances for which existence has not yet been determined.  The Constructive Protocol CPro1 uses LLMs to generate search heuristics that have the potential to construct solutions to small open instances.  Starting with a textual definition and a validity verifier for a particular type of design, CPro1 guides LLMs to select and implement strategies, while providing automated hyperparameter tuning and execution feedback.  CPro1 with reasoning LLMs successfully solves long-standing open instances for 7 of 16 combinatorial design problems selected from the 2006 \textit{Handbook of Combinatorial Designs}, including new solved instances for 3 of these (Bhaskar Rao Designs, Symmetric Weighing Matrices, Balanced Ternary Designs) that were unsolved by CPro1 with non-reasoning LLMs.  It also solves open instances for several problems from recent (2025) literature, generating new Covering Sequences, Johnson Clique Covers, Deletion Codes, and a Uniform Nested Steiner Quadruple System. 

\end{abstract}

\section{Introduction}
\label{sec:introduction}
We apply code generation via reasoning Large Language Models (LLMs) to a specific task in the mathematical field of combinatorial design.  \textit{Combinatorial designs} are systems of finite sets that satisfy specified constraints.  The particular finite sets and constraints involved define the {\em type} of combinatorial design (e.g. Balanced Incomplete Block Design, Packing Array).  The \textit{existence problem} (or \textit{combinatorial design problem}) for a particular type of combinatorial design has a small number of input \textit{parameters} (e.g. size), and asks whether it is possible to construct a design that satisfies the constraints for these parameters.  An \textit{instance} of the existence problem specifies particular numerical values for the parameters.  The existence problem is often addressed by systematic mathematical constructions that are proven correct and show existence for large classes of instances, and by mathematical proofs of impossibility that show certain instances cannot exist.  But these may leave behind parameters for which existence remains unknown; these remain as open instances. 

As an example, a \textit{Symmetric Weighing Matrix} is an $n \times n$ matrix $W$ with entries in the set $\{0,1,-1\}$, satisfying $W W^T=wI$ and $W=W^T$.  For $w=16$, a Symmetric Weighing Matrix exists for $n=16$, $n=18$, and all $n \geq 20$ with the possible exception of $n \in \{22,23,25,27,29\}$ for which existence was unknown \cite{handbook} as of 2006.  A 2023 result showed that $w=16$ $n=23$ exists \cite{georgiou}.  This paper's results include construction of a $w=16$ $n=22$ Symmetric Weighing Matrix, resolving this open instance and leaving $n \in \{25,27,29\}$ as the remaining open instances for $w=16$.

One standard approach to small open instances is heuristic computational search (\textit{Handbook of Combinatorial Designs} \cite{handbook}, chapter VII.6).  This paper builds on earlier work which developed a Constructive Protocol \textit{CPro1} that automates an experimental process to identify and optimize heuristic strategies \cite{jan2025preprint}.  Starting from a textual definition for a combinatorial design problem and a validity verifier for proposed solutions, CPro1 guides LLMs to select and implement strategies in C code, while providing automated hyperparameter tuning and execution feedback using the verifier (Algorithm~\ref{alg:cpro}).  In this paper, we use CPro1 with \textit{reasoning} LLMs that are trained with a reinforcement learning process to develop their answers in a lengthy textual process that allows space for iteration and revision before finalizing the answer \cite{r1}.   

We assess the ability of CPro1, with a reasoning LLM, to solve open instances of combinatorial design problems.  CPro1 successfully solves long-standing open instances for 7 of 16 combinatorial design problems selected from the 2006 \textit{Handbook of Combinatorial Designs}; 3 of these 7 with new open instances solved compared to CPro1 using non-reasoning models \cite{jan2025preprint}.  We also find CPro1 solves open instances for 3 out of 4 problems from recent (February 2025) combinatorial design literature.  CPro1 also improves upon an April 2025 results that used the FunSearch LLM-based protocol to create novel Deletion Codes \cite{deletioncodes}.  Positive results are shown in Table~\ref{tab:mainresults}.  While we don't see success with combinatorial design problems that have already seen sustained iterative development of computational methods by the research community (e.g. Covering Arrays \cite{covsa1,torres-jimenez-ieee-access-2019,jcdcoveringarrays}), we do see strong results in design types that have seen less attention on computational methods.

\begin{algorithm*}[h!]
    \caption{Protocol \textbf{CPro}.  Note \textbf{prompt}(x) returns LLM result when prompted with x.}
    \label{alg:cpro}
    \phantom{ini}\textbf{Input}: Definition (of the problem), Dev \& Open (instance parameter lists), Verifier (Python code) \\
    \phantom{ini}\textbf{Output}: Verified designs for the Open instances (if found)\\
    \vspace*{-4pt}
    \begin{algorithmic} 
        \STATE{// \textbf{Generate 1000 diverse candidate search heuristics, using the C programming language}} \\
        \FOR{50 reps}
        \STATE Strategies = \textbf{prompt}(Definition + "Please suggest " + N(=20) + " different approaches...") 
        \FOR{S in Strategies}
        \STATE Details = \textbf{prompt}(Definition + "We have selected..." + S + "...Describe the elements...") 
        \STATE Code,Hyperparm\_ranges = \textbf{prompt}("Now implement...") //Continued chat; Details in context\\
        \STATE \textbf{append} Code,Hyperparm\_ranges to Candidates
        \ENDFOR
        \ENDFOR \phantom{x} // We now have R*N=1000 candidates\vspace{4pt}
        \STATE{// \textbf{Execute each candidate (up to 50 sec.), tune hyperparameters, use Verifier to score results}}
        \FOR{C in Candidates}
        \STATE \textbf{Set} C's Hyper\_settings,Score = hypertune(C,Hyperparm\_ranges) // Execute on Dev instances \\
        // Grid-based hyperparm tuning (linear in middle, logarithmic/fine-grained out to the endpoints). \\
        // Up to 1000 grid points run for 0.5 seconds; top 100 for 5 sec.; top 10 for 50 sec.; return best 1. \\
        \ENDFOR
        \STATE \textbf{truncate} Candidates to the top 5 according to their Score\vspace{4pt}
        \STATE{// \textbf{Optimize the execution speed of the top 5 candidates}} \\
        \FOR{C in Candidates}
        \FOR{5 optimization rounds}
        \STATE O = [\textbf{prompt}("...improve the performance..." + C + "...") * 50 times] \\
        \STATE \textbf{if} highest-scoring result from O is much better than C: \textbf{replace} C by it, \textbf{else}: break \\
        \ENDFOR
        \ENDFOR\vspace{4pt}
        \STATE{// \textbf{Execute top 5 candidates for a full 2 hours on Dev instances and verify results}} \\
        \STATE \textbf{for} C in Candidates \textbf{do}: score C with 2-hour runs on Dev instances
        \STATE \textbf{truncate} Candidates to the top 2 according to their 2-hour score\vspace{4pt}
        \STATE{// \textbf{Execute top 2 candidates for 48 hours on Open instances and output verified solutions}} \\
        \FOR{C in Candidates}
        \STATE Run C for 48 hours on Open instances, and \textbf{output} each result that passes Verifier check.
        \ENDFOR        
        
    \end{algorithmic}
\end{algorithm*}

\begin{table}
    \centering
    \begin{subtable}{\linewidth}
    \centering
    \begin{tabular}{| m{0.97\linewidth}| }
    \hline
       \vspace*{1pt} \textit{Packing \mbox{Array} (PA)}:  an $N \times k$ array with entries in the set $\{0,1,\ldots v-1\}$, such that every $N\times 2$ subarray contains every ordered pair of symbols at most once.\\
       \rowcolor{green!5}  \begin{tabular}{l l l l l l l}
        (N,k,v): & \textit{(24,7,6)} & \textit{(18,8,6)} & \textit{(28,10,8)} & \textit{(24,11,8)} & \textit{(32,11,9)} &
        \textit{(28,12,9)} \\
        \textit{(21,14,9)} & \textit{(19,15,9)}
        \end{tabular}
        \\
        \hline
        \vspace*{1pt} \textit{Symmetric \mbox{Weighing} Matrix (SymmW)}: an $n \times n$ matrix $W$ with entries in the set $\{0,1,-1\}$, satisfying $W W^T = wI$ and $W = W^T$.  \\
        \rowcolor{green!5} \begin{tabular}{l l l l}
         (n,w): & \textit{(19,9)}$\dagger$ & \textit{(21,9)}$\dagger$ & \textbf{\textit{(22,16)}}
        \end{tabular}
        \\
        \hline
         \vspace*{1pt} \textit{Skew \mbox{Weighing} \mbox{Matrix} (SkewW)}: an $n \times n$ matrix $W$ with entries in the set $\{0,1,-1\}$, satisfying $W W^T = wI$ and $W^T=-W$. \\
         \rowcolor{green!5} \begin{tabular}{l l}
        (n,w): &\textit{(18,9)} 
        \end{tabular}
        \\
        \hline
        \vspace*{1pt} \textit{Bhaskar Rao Design (BRD)}: a $v \times b$ array with entries in the set $\{-1,0,1\}$.  Each row contains $r$ nonzero entries and each column contains $k$ nonzero entries.  For any pair of distinct rows, the pairwise element products contain $-1$ and $+1$ each $L/2$ times. \\
       \rowcolor{green!5} \begin{tabular}{l l l l}
        (v,b,r,k,L): & \textbf{\textit{(15,42,14,5,4)}} & \textbf{\textit{(15,126,42,5,12)}} &  \textbf{\textit{(16,48,15,5,4)}} 
        \end{tabular}
        \\
        \hline
          \vspace*{1pt} \textit{Balanced Ternary Design (BTD)}: an arrangement of $V$ elements into $B$ multisets, or blocks, each of cardinality $K \leq V$, satisfying: 1. Each element appears $R=p1 + 2*p2$ times, with multiplicity one in $p1$ blocks and multiplicity two in $p2$ blocks. 2. Every pair of distinct elements appears $L$ times. \\
        \rowcolor{green!5}  \begin{tabular}{l l l l}
        (V,B;p1,p2,R;K,L): & \textit{(17,17;8,2,12;12,8)} & \textit{(14,21;6,3,12;8,6)} & \textit{(12,16;4,4,12;9,8)} \\
        \textit{(16,16;7,3,13;13,10)} & \textit{(12,21;4,5,14;8,8)} & \textbf{\textit{(16,22;9,1,11;8,5)}} & \textbf{\textit{(21,21;12,1,14;14,9)}}
        \end{tabular}
        \\
        \hline
        \vspace*{1pt} \textit{Equidistant \mbox{Permutation} \mbox{Array}~(EPA)}: $m$ rows, each a permutation of $\{0,1,\ldots n-1\}$.  Each pair of distinct rows must differ in exactly $d$ positions.\\
        \rowcolor{green!5} \begin{tabular}{l l}
         (n,d,m): & \textit{(12,8,21)}$\dagger$ 
        \end{tabular}
        \\
        \hline
        \vspace*{1pt} \textit{Florentine \mbox{Rectangle} (FR)}: $r$ rows, each a permutation of $S=\{0,1,\ldots n-1\}$. For distinct $a,b \in S$ and $m\in \{1,2,...,n-1\}$, at most one row has $b$ positioned $m$ steps to the right of $a$. \\
        \rowcolor{green!5} \begin{tabular}{l l l l l l}
        (r,n): & \textit{(7,20)} & \textit{(7,24)} & \textit{(7,25)} & \textit{(7,26)} & \textit{(7,27)} 
        \end{tabular}
        \\
        \hline
    \end{tabular}
    \caption{Open instances from 2006 \textit{Handbook}, that were solved and found to exist by CPro1. ($\dagger$: in prototyping set)}
    \end{subtable}
   
\begin{subtable}{\linewidth}
\centering
    \begin{tabular}{| m{0.97\linewidth} |}
    \hline
    \vspace*{1pt} \textit{Covering Sequence (CS)}: A cyclic sequence $x_0,x_1, \ldots x_{L-1}$ of length $L$ over the binary alphabet, such that for any length-$n$ binary word there exists a $j$ such that subsequence $x_j,x_{(j+1) \bmod{L}},x_{(j+2) \bmod{L}}, \ldots x_{(j+n-1) \bmod{L}}$ is Hamming distance at most $R$ from the word.\\
    \rowcolor{green!5} \begin{tabular}{l l l l l l}
   (n,R,L): & \textbf{\textit{(9,1,71)}} & \textbf{\textit{(10,1,138)}} & \textbf{\textit{(11,1,224)}} & \textbf{\textit{(11,2,64)}} & \textbf{\textit{(12,2,127)}} \\
    \textbf{\textit{(12,3,36)}} & \textbf{\textit{(13,2,276)}} & \textbf{\textit{(13,3,61)}} & \textbf{\textit{(14,3,122)}} & \textbf{\textit{(15,3,230)}} &  \textbf{\textit{(16,3,426)}}\\
    \textbf{\textit{(17,2,3938)}} & \textbf{\textit{(17,3,795)}} & \textbf{\textit{(18,1,52390)}} & \textbf{\textit{(18,2,7605)}} & \textbf{\textit{(18,3,1481)}} &  \textbf{\textit{(19,1,104498)}}\\
    \textbf{\textit{(19,2,14797)}} &  \textbf{\textit{(19,3,2734)}} & \textbf{\textit{(20,1,207000)}} & \textbf{\textit{(20,2,28901)}} &  \textbf{\textit{(20,3,5102)}}
    \end{tabular}
    \\
    \hline
    \vspace*{1pt} \textit{Johnson Clique Cover (JCC)}: The Johnson Graph $J(N,k)$ has a vertex for each $k$-element subset of $\{1,2,\ldots N\}$.  Two subsets $A,B$ are connected by an edge if $|A \cap B| = k-1$.  A size-$C$ Johnson Clique Cover has $C$ cliques in $J(N,k)$ such that their union includes all vertices in $J(N,k)$.\\
    \rowcolor{green!5} \begin{tabular}{l l l l l l}
    (N,k,C): & \textbf{\textit{(13,4,105)}} &  \textbf{\textit{(13,6,248)}} &  \textbf{\textit{(14,4,138)}} & \textbf{\textit{(14,6,410)}} & \textbf{\textit{(15,4,177)}}
    \end{tabular}
    \\
     \hline
    \vspace*{1pt} \textit{Uniform Nested Steiner Quadruple System (UNSQS):} A \textit{Steiner Quadruple System (SQS)} is a set of blocks, each a size-$4$ subset of $V=\{0,1,...,v-1\}$, such that each subset of $3$ elements of $V$ is contained in exactly one block. A \textit{Uniform Nested SQS} splits each block into two \textit{ND-pairs} of $2$ elements each, such that each distinct ND-pair appears the same number of times.  Let $p$ denote the number of distinct ND-pairs that appear among the blocks. \\
    \rowcolor{green!5} \begin{tabular}{l l}
    (v,p): & \textbf{\textit{(14,91)}} 
    \end{tabular}
    \\
     \hline
     \end{tabular}     
    \caption{Open instances from Feb. 2025 articles \cite{covseq,jclique,unsqs}, that were solved and found to exist by CPro1.}
\end{subtable}

\begin{subtable}{\linewidth}
\centering
    \begin{tabular}{| m{0.97\linewidth}| }
    \hline
    \vspace*{1pt} \textit{Deletion Code (DC)}: a set of $m$ binary words of length $n$.  Given word $x$, $D(x)$ is the length-($n-s$) words obtained by deleting $s$ distinct bits.  For any two distinct words $x,y$: $|D(x) \cap D(y)|=0$.\\
    \rowcolor{green!5}  \begin{tabular}{l l l l l l}
    (n,s,m): & \textbf{\textit{(12,2,36)}} & \textbf{\textit{(13,2,55)}} & \textbf{\textit{(14,2,85)}} & \textbf{\textit{(15,2,132)}} & \textbf{\textit{(16,2,208)}} \\
    \textbf{\textit{(13,3,16)}} & \textbf{\textit{(14,3,21)}} & \textbf{\textit{(15,3,29)}}  & \textbf{\textit{(16,3,42)}}
    \end{tabular}
    \\
     \hline
         \end{tabular}
    \caption{Deletion Codes found to exist by CPro1, improving on results from Apr. 2025 article using FunSearch \cite{deletioncodes}.}
    \end{subtable}
    \caption{\textbf{Main Results.} Combinatorial design problems by source, each with brief definition and a list of the open instances solved (bold italic indicates newly solved instances using CPro1 with a reasoning model; plain italic previously solved with CPro1 and non-reasoning models \cite{jan2025preprint}). For each of these open instances, the code generated by CPro1 constructed a verified solution. The Appendix includes sample solutions (Figs.~\ref{fig:arrays1} to \ref{fig:arrays7}) and full definitions as used by CPro1 (Tables~\ref{tab:definitions1} to \ref{tab:definitions4}).}
    \label{tab:mainresults}
\end{table}

\section{Related Work}

Code generation is one of the primary applications of LLMs \cite{codegen,codegen2,llmsforscience}.  LLM code generation has been used to develop heuristics for combinatorial optimization problems \cite{eoh,ceoh,cobench}, including generating search operators for genetic algorithms \cite{reevo}.  In this paper, we use LLMs to propose and implement heuristic strategies in an open-ended way, and successful strategies that emerge include genetic algorithms \cite{ga}, simulated annealing \cite{sa}, and tabu search \cite{ts}.
 
Efforts to apply LLMs to mathematics have focused on exercises and benchmarks with known solutions \cite{math,olympiad,frontiermath,streamlining,combibench} and generation of step-by-step proofs that could be verified with systems like Lean \cite{leanllm,oracleverifier}.  LLMs have difficulty generating valid lengthy step-by-step proofs \cite{matharena}.  Here, we focus on problems that can be resolved by constructing a combinatorial object that can be easily verified, rather than requiring a step-by-step proof.  This has the potential to resolve open questions in mathematics without the difficulty of generating lengthy step-by-step proofs.

This paper builds on earlier work which developed Constructive Protocol CPro1 \cite{jan2025preprint} for combinatorial design problems using non-reasoning LLMs. Here, we focus on the use of \textit{reasoning} LLMs that are generally more effective than non-reasoning LLMs for mathematics and code generation \cite{r1,o3mini,livebenchpaper,livebenchwebsite,frontiermath,frontiermathwebsite}.

Deep learning has emerging applications to constructions in research-level mathematics \cite{wagneroriginal,wagnerextremal,patternboost,polytopes,conjecturing,alphaevolve}.  \textit{FunSearch} \cite{funsearch} uses code-generating LLMs in an evolutionary algorithm to search for greedy functions.  It succeeded on a recognized open question by constructing a size-512 \textit{Cap Set} for $n=8$, and was recently used to obtain novel \textit{Deletion Codes} \cite{deletioncodes}.  Compared to FunSearch, CPro1 generates fewer candidates, allows open-ended strategies rather than just greedy functions, and is simpler without FunSearch's iterative evolution.  We test CPro1 with a reasoning model on Cap Sets and Deletion Codes.

\section{Method}
\label{sec:method}

\textbf{Terminology}: A \textit{Large Language Model} (LLM) takes a textual \textit{prompt} and returns a textual response, which may include natural language and/or programming language code.  LLMs are trained via machine learning, but we use off-the-shelf pretrained LLMs.  For \textit{reasoning} LLMs, the pretraining includes reinforcement learning of a textual process that iteratively develops and refines the answer; this textual reasoning trace is not part of the final result.  We use \textit{protocol} to refer to an algorithm which includes calls to an LLM, and \textit{scaffolding} consists of protocol elements other than the LLM.

\begin{table*}[tbh]
\centering
\begin{minipage}{0.54\textwidth}
\centering
{\rowcolors{3}{black!8}{black!1}
\begin{tabular}{m{5.5cm} m{1.3cm}}
\textbf{Selected combinatorial designs from the \textit{Handbook}} & \textit{Handbook} Chapter  \\
\hline
Balanced Incomplete Block Design & II.1   \\
Packing Array (PA) & III.3    \\
Orthogonal Array & III.6 \\
Symmetric Weighing Matrix (SymmW) & V.2  \\
Skew Weighing Matrix (SkewW) & V.2  \\
Bhaskar Rao Design (BRD) & V.4  \\
Balanced Ternary Design (BTD) & VI.2\\
Costas Array & VI.9  \\
Covering Design & VI.11 \\
Difference Triangle Set &  VI.19 \\
Perfect~\mbox{Mendelsohn}~\mbox{Design} & VI.35  \\
Equidistant Permutation Array (EPA) & VI.44  \\
Florentine Rectangle (FR) & VI.62   \\
Circular Florentine Rectangle & VI.62   \\
Tuscan-2 Square & VI.62  \\
Supersimple Design & VI.57 \\
\end{tabular}
}
\vspace{-0.27em}
\end{minipage}
\hfill
\begin{minipage}{0.42\textwidth}
\centering
{\rowcolors{3}{black!8}{black!1}
\begin{tabular}{m{3.6cm} m{1.4cm}}
\textbf{Combinatorial designs from Feb. 2025} & \phantom{Source} Source  \\
\hline
Covering Sequence & \cite{covseq} \\
Johnson Clique Cover & \cite{jclique} \\
Uniform Nested Steiner Quadruple System & \cite{unsqs} \\
Covering Array & \cite{jcdcoveringarrays} \\
\hline
\phantom{placeholder}  &  \\
\end{tabular}
}
{\rowcolors{3}{black!8}{black!1}
\begin{tabular}{m{3.6cm} m{1.4cm}}
\textbf{FunSearch Problems} & Source  \\
\hline
Cap Sets & \cite{funsearch} \\
Deletion Codes & \cite{deletioncodes} \\
\hline
\end{tabular}
}
\caption{\textbf{Problem Selection}: Combinatorial designs with long-standing open instances from the 2006 \textit{Handbook of Combinatorial Designs} \cite{handbook}; recent problems from Feb. 2025 combinatorial design literature; 2 problems addressed by FunSearch.}
\label{tab:selected}
\end{minipage}
\rule{\linewidth}{0.4pt}
\end{table*}

\subsection{Selection of Combinatorial Designs (see Table~\ref{tab:selected})}
\label{sec:selection}
We start with the same 16 types of combinatorial designs from the 2006 \textit{Handbook of Combinatorial Designs} \cite{handbook} (henceforth \textit{Handbook}) that were used with CPro1 and non-reasoning models \cite{jan2025preprint}.  Each of these has clearly defined open instances in the \textit{Handbook} with relatively small parameters that might be amenable to heuristic search.  Instances which have already been solved in the literature are omitted \cite{stardom,packingsat,designs2002,georgiou,dinitzupdate5,lajollacovering,lastpmd,dinitzupdate6}, focusing on remaining instances that are still open.  

In addition, we also sample much more recent open questions.  From the \textit{Journal of Combinatorial Design} and arXiv's math.CO combinatorics category, we survey all articles that first appeared in Feb. 2025, and select those which describe combinatorial design problems and include a table of specific open instances.  Out of 6 \textit{J. Combinatorial Design} articles, we select 2, and out of 447 arXiv combinatorics articles (the great majority of which do not address combinatorial designs), we select 2.  Note this gives a rough idea of the scope of applicability of the method. 

We also include Cap Sets from the original FunSearch paper \cite{funsearch} and Deletion Codes from an April 2025 FunSearch paper \cite{deletioncodes}, to see if CPro1 could replicate FunSearch's results and go beyond them.

For each selected problem, we provide:
\begin{description}[leftmargin=0pt]
\item[Textual Definition] of the problem, mandating a specific solution representation as an integer array.
\item[Verifier in Python] Determines whether a proposed solution in this representation is correct.
\item[Open Instances] Instance parameters for which existence is not yet known.  The ultimate goal is to construct designs with these parameters.
\item[Development Instances (Dev Instances)] Instances known to exist, including some of the smallest, as well as ones just slightly smaller than the Open Instances.  Candidates are executed on these, with results checked by the Verifier, to identify the most promising approaches.
\end{description}

\subsection{Protocol}
\label{sec:methoddev}

Algorithm~\ref{alg:cpro} summarizes the Constructive Protocol CPro1 (see \cite{jan2025preprint} for more details).  During initial testing, 1000 candidate programs are generated and undergo hyperparameter tuning, and each are scored on the basis of 50 seconds of execution using the tuned hyperparameters.  The 5 top-scoring candidates proceed to optimization (also 50 seconds).  For final testing on development instances, the same 5 candidates (after optimization) are given 2 hours.  The 2 top-scoring candidates from this then run for 48 hours on the open instances.  

\subsection{Prototyping Set}
CPro1 was developed with the aid of a small {\em prototyping set} of combinatorial design problems for which a manual effort found solvable open instances \cite{jan2025preprint}.  This was based on manual development and tuning of local search methods for 5 of of the 16 selected combinatorial designs from the \textit{Handbook}: Bhaskar Rao Designs (BRD), Difference Triangle Sets, Equidistant Permutation Arrays (EPA), Supersimple Designs, and Symmetric Weighing Matrices (SymmW).  The local search methods selected changes which minimize a cost function, while sometimes accepting worsening moves to escape local optima.  Local search succeeded in solving 1 open instance for EPA and 2 for SymmW (marked with $\dagger$ in Table~\ref{tab:mainresults}(a)); EPA and SymmW formed the Prototyping Set.

Table~\ref{tab:models} shows CPro1's Prototyping Set results with the non-reasoning GPT-4o model used originally, as well as with two reasoning models: the OpenAI o3-mini \cite{o3mini} set to “high” reasoning, and the open-weights model DeepSeek R1 \cite{r1}.  o3-mini-high is especially successful and solves 4 instances from the prototyping set (including SymmW instance $n=22$ $w=16$ that both hand-coded local search and CPro1 with GPT-4o failed on), so we use o3-mini-high for reasoning model experiments here.

\begin{table}[t]
\centering
 {\rowcolors{3}{black!8}{black!1}
\begin{tabular}{m{3.8cm} m{1.65cm}  m{2.95cm} c}
LLM &  Ver. Date &  \centering SymmW open solved & EPA open solved \\
\hline
GPT-4o \cite{gpt4o} & 2024-05-13 & \centering 2 &  1 \\
o3-mini-high \cite{o3mini} & 2025-01-31 & \centering 3 & 1 \\
DeepSeek R1 \cite{r1} & 2025-01-20 & \centering 2 & 1 \\
\hline
\end{tabular}
}
\vspace*{4pt}

\caption{\textbf{Prototyping Set Results.} For each LLM and total number of candidate programs: the number of distinct open instances that are solved by CPro1 on each Prototyping Set problem.  GPT-4o is a non-reasoning model previously used with CPro1 \cite{jan2025preprint}; o3-mini-high and DeepSeek R1 are reasoning models newly tested here.}
\label{tab:models}
\end{table}

\subsection{Experiments}
\label{sec:experiments}

For each of the 16 types of combinatorial design from the \textit{Handbook}, each of the four Feb. 2025 problems, and both of the FunSearch problems, we run CPro1 with the reasoning model o3-mini-high.  We also include and compare prior results with obtained with CPro1 using non-reasoning model GPT-4o \cite{jan2025preprint} on the \textit{Handbook} problems.  We also test ablated and scaled-down versions of the protocol, using the same candidate programs generated by the original runs.   

When we succeed in solving open instances of a combinatorial design problem, we extend by testing the generated code on adjacent open instances (e.g. next size larger).

Each experiment runs on a Linux machine with AMD Ryzen 9 7950X3D CPU and 128GB of memory, with the machine fully dedicated to one run at a time.  A full run of CPro1 on one type of combinatorial design takes approximately 6-10 days, the majority of which is used running candidate programs on development instances and open instances.  Runs with o3-mini-high tend to take longer than runs with GPT-4o, because with o3-mini-high there are fewer candidate programs which immediately fail.

\section{Results}
\label{sec:results}

Table~\ref{tab:mainresults} shows the main results.  CPro1 solves open instances for 7 types of combinatorial designs of the 16 selected from the \textit{Handbook}.  This includes open instances that were only solved with CPro1 using the reasoning model o3-mini-high (and not using GPT-4o) for 3 of these 7.  Positive results using the reasoning model include Bhaskar Rao Designs, for which our hand-coded local search failed.  CPro1 with o3-mini-high solves a superset of instances solved with GPT-4o (Table~\ref{tab:successes}).  For other \textit{Handbook} combinatorial designs from Table~\ref{tab:selected}, code generated by CPro1 solves many development instances, but no open instances.

CPro1 using o3-mini-high solves open instances for 3 of the 4 selected problems from Feb. 2025 combinatorial design literature.  It creates problem-specific local search algorithms for Covering Sequence and Johnson Clique Cover, obtaining significant progress on open instances compared to the computational methods used in the original publications \cite{covseq,jclique}.  These are more recent types of combinatorial designs that do not appear in the \textit{Handbook}, and have thus far received less attention.   

For the Deletion Codes problem, FunSearch was reported in April 2025 to improve on state-of-the-art codes for 3 sets of parameters \cite{deletioncodes}.  CPro1 implements a tabu search which replicates these results and also finds further improvement (larger sets of codewords meeting the constraints) for all 3, plus 6 other sets of parameters, substantially improving on the state of the art for small Deletion Codes.

Note Table~\ref{tab:mainresults} only shows independent solved instances.  For example, the EPA with n=12 d=8 m=21 implies the existence of n=13 d=8 m=21 (add a constant column), and the PA with N=32 k=11 v=9 trivially implies the existence of N=31 k=11 v=9 (remove a row), but we don't list these. 

All of the open instance solutions, and the code that constructed them, are available on github. \footnote{https://github.com/Constructive-Codes/CPro1}

\begin{table}[t]
\begin{tabular}{|m{2.5cm}|m{1.13cm}|m{1.13cm}|m{1.13cm}|m{1.13cm}|m{1.13cm}|m{1.13cm}|m{1.13cm}|}
\hline
 & PA & SymmW & SkewW & BTD & FR & EPA & BRD \\
\hline
\textbf{o3-mini-high} & \cellcolor{lime}Tabu &\cellcolor{lime} DFS & \cellcolor{lime} DFS & \cellcolor{lime} Tabu& \cellcolor{lime} GRASP & \cellcolor{lime}cSA & \cellcolor{lime} cSA,2p \\
\hline
- Reduce runtime  & \cellcolor{lime}Tabu & \cellcolor{lime} DFS & \cellcolor{lime} DFS  & \cellcolor{lime} Tabu &   & \cellcolor{lime}cSA & \cellcolor{lime} cSA,2p \\
\hline
- No final dev test & \cellcolor{lime}Tabu  & \cellcolor{lime} DFS  &  \cellcolor{lime} DFS  & \cellcolor{lime} Tabu  &   &  & \cellcolor{lime} cSA,2p \\
\hline
- No optimization & \cellcolor{lime}Tabu & \cellcolor{lime} DFS & \cellcolor{lime} DFS & \cellcolor{lime} Tabu  &  & & \cellcolor{yellow!50} 2p \\
\hline
- No hyper tuning   & \cellcolor{lime}Tabu & \cellcolor{lime} DFS & \cellcolor{lime} DFS & \cellcolor{lime} Tabu,RG & & & \cellcolor{yellow!50} 2p \\
\hline
\hline
\textbf{GPT-4o} & \cellcolor{lime}cSA &\cellcolor{lime} rSA & \cellcolor{lime} cSA & \cellcolor{lime} GA& \cellcolor{lime} DFS & \cellcolor{lime}cSA & \\
\hline
- Reduce runtime  & \cellcolor{lime}cSA & \cellcolor{lime} rSA & \cellcolor{lime} cSA  & \cellcolor{yellow!50} GA & \cellcolor{yellow!50} DFS  & & \\
\hline
- No final dev test & \cellcolor{lime}cSA  & \cellcolor{lime} rSA  &  \cellcolor{lime} cSA  & \cellcolor{yellow!50} GA  & \cellcolor{yellow!50} DFS  & &  \\
\hline
- No optimization & \cellcolor{lime}cSA & \cellcolor{yellow!50} SA & \cellcolor{lime} cSA & \cellcolor{yellow!50} GA  & & &  \\
\hline
- No hyper tuning   & \cellcolor{yellow!50}rSA & \cellcolor{lime}GA & & & & &  \\
\hline
\end{tabular}
\vspace{3pt}
\caption{\textbf{Problem-Specific Results and Ablation.} Each row is an experiment, each column a type of combinatorial design from the \textit{Handbook}.  Colored cells show where the experiment solved at least one open instance: green indicates the experiment solved all the open instances that the initial run did, and yellow indicates an ablation experiment solved only a subset.  The first set of rows use o3-mini-high, and the ``-'' rows stack up successive ablations: \textbf{Reduce runtime} reduces from 48 hours to 2 for open instances, \textbf{No final dev test} eliminates final 2 hour testing on development instances (instead using original 50-second test results), \textbf{No optimization} eliminates the code optimization step, and \textbf{No hyper tuning} eliminates hyperparameter tuning (instead using defaults given by the LLM).  The second set of rows use GPT-4o (from~\cite{jan2025preprint}).  Each cell shows the strategies that obtained success: \textbf{DFS}: backtracking depth-first search, \textbf{GA}: genetic algorithm, \textbf{SA}: simulated annealing with slow cooling schedule, \textbf{rSA}: simulated annealing with periodic resets, \textbf{cSA}: constant-temperature simulated annealing, \textbf{Tabu}: Tabu search \cite{ts}, \textbf{RG}: randomized greedy (randomized DFS within row, restart if any row fails), \textbf{2-phase} for BRD: find binary incidence matrix, then add signs, \textbf{GRASP:} Greedy randomized adaptive search procedure \cite{GRASP}.}
\label{tab:successes}
\end{table}

\subsection{Ablation}
Table~\ref{tab:successes} shows that all considered elements of the protocol are needed to obtain the full results on the 7 successful \textit{Handbook} problems, though CPro1 runs with o3-mini-high are less degraded by ablation.

\subsection{Scaled-Down Runs}
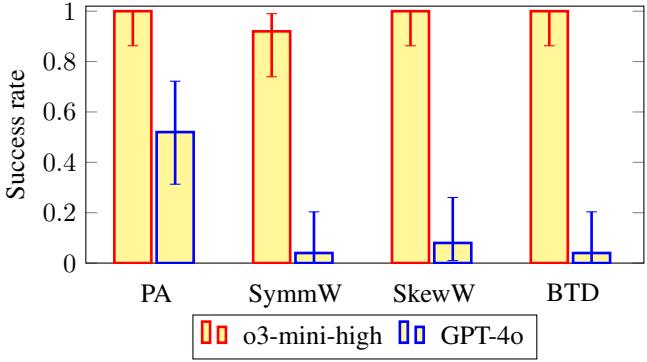
\begin{figure}[t]
\centering
\begin{minipage}[t]{0.6\textwidth}
\begin{tikzpicture}[baseline=(current bounding box.north)]
\begin{axis}[
    width=9cm,
    height=5cm,
    ybar,
    bar width=14pt,
    enlarge x limits=0.17,
    ymin=0, ymax=1.02,
    ylabel={Success rate},
    symbolic x coords={PA, SymmW, SkewW, BTD},
    xtick={PA, SymmW, SkewW, BTD}, 
    xtick style={draw=none},
    legend style = {
      at={(0.5,-0.2)}, anchor=north,
      legend columns = 2,
      column sep = 4pt,
      align = left
    },
    bar shift=-9pt,
    error bars/.cd,
]

\addplot+[
    draw=red,
    line width=1pt,
    fill=yellow!50,
    error bars/.cd,
        y explicit,
        y dir=both,
        error bar style={red,line width=1pt}
] table[
    x=Problem,
    y=Rate,
    y error plus=Yplus,
    y error minus=Yminus
]{
Problem  Rate  Yplus   Yminus
PA       1.000 0.0000  0.1372
SymmW    0.920 0.0702  0.1803
SkewW    1.000 0.0000  0.1372
BTD      1.000 0.0000  0.1372

};
\addlegendentry{o3-mini-high}

\addplot+[
    bar shift=+7pt,
    draw=blue,
    line width=1pt,
    fill=yellow!50,
    error bars/.cd,
        y explicit,
        y dir=both,
        error bar style={blue,line width=1pt}
] table[
    x=Problem,
    y=Rate,
    y error plus=Yplus,
    y error minus=Yminus
]{
Problem  Rate  Yplus   Yminus
PA       0.520 0.2020  0.2069
SymmW    0.040 0.1635  0.0390
SkewW    0.080 0.1803  0.0702
BTD      0.040 0.1635  0.0390

};
\addlegendentry{GPT-4o}

\end{axis}
\end{tikzpicture}
\end{minipage}%
\hfill
\begin{minipage}[t]{0.35\textwidth}
\vspace{0.5em}
\caption{\textbf{Results from scaled-down 40-candidate runs.} With reduced runtime (2 hours) on open instances, no final dev test, and no optimization.  Each bar shows the rate of successfully solving at least one open instance, across 25 repeated runs, with 95\% Clopper-Pearson confidence interval.  Each problem's difference between o3-mini-high and GPT-4o is significant (p<0.0001, Z-test).} 
\label{fig:downscaled}
\end{minipage}
\end{figure}

Figure~\ref{fig:downscaled} has results from repeated scaled-down runs for 4 of the \textit{Handbook} problems, each using only 40 candidate programs.  These candidates come from the original pool of candidates, giving $1000/40=25$ scaled-down runs.  Each bar shows the fraction of these 25 runs that solved at least one open instance.  These are ablated runs: using reduced runtime of 2 hours on open instances, no final dev test, and no optimization step (but retaining hyperparameter tuning).  We see that success rates remain high for the o3-mini-high reasoning model, whereas GPT-4o performs relatively poorly.

\subsection{Strategies Implemented by Generated Code}
\label{sec:strategies}

Most of the positive results in Table~\ref{tab:successes} use simulated annealing, genetic algorithms, tabu search, or depth-first search.  Each full run includes 1000 candidates, generated from 50 lists of 20 proposed strategies each.  Simulated annealing, genetic algorithms, tabu search, greedy algorithms, and depth-first search are usually each proposed dozens of times during a run.  This gives adequate room to explore alternatives within each method (e.g. for cost function, neighborhood).

All of the successful strategies are randomized, and most appear well-optimized. Successful programs are 120-270 lines of C code for GPT-4o, and up to 540 lines for o3-mini-high.

For the Uniform Nested Steiner Quadruple System problem, CPro1's approach succeeds by addressing different constraints in two phases: first create a Steiner Quadruple System using Knuth's Algorithm X \cite{algorithmx}, then perform tabu search for nested pairs.  A two-phase approach also arises for Bhaskar Rao Designs: first find a 0/1 matrix that meets incidence constraints, then add +1/-1 signs.  These two-phase approaches seem somewhat surprising; one might have anticipated that constraint interaction would prevent success.  Prior two-phase approaches for the Covering Array combinatorial design problem \cite{covsa2,torres-jimenez-ieee-access-2019} have both phases addressing the same constraints.
 
\section{Limitations}
\label{sec:limitations}

In general, the application of simulated annealing, tabu search, and genetic algorithms to combinatorial designs is well established (\textit{Handbook} chapter VII.6).  The positive results reported here arise from automation of computational experimentation that could have been done manually.  The research community has only undertaken limited effort on such experimentation for the designs with positive results here; this may have left low-hanging fruit for CPro1.  Some of the designs with no positive results here have received much greater attention from the research community.  For example, CPro1 was unsuccessful on Covering Arrays.  This included failure to replicate mathematical constructions from the Feb. 2025 paper, as well as earlier results from specialized search heuristics \cite{covsa1,covsa2} that were noted in the Feb. 2025 paper.  Covering Arrays have received substantial attention, leading to development of progressively better computational techniques and mathematical constructions \cite{covsa1,torres-jimenez-ieee-access-2019}; CPro1 is less likely to contribute here.

CPro1 can only show existence by constructing a solution.  For some of the open instances included here, there may exist no solution, but non-existence proofs are outside the scope of CPro1.

We perform only one full-scale run of CPro1 with each included LLM on each type of combinatorial design, and the LLMs are inherently nondeterministic in their responses; repeat runs could yield different results.

After our experiments were completed, a new publication reported use of FPGAs to solve some of the open Difference Triangle Set instances that CPro1 failed to solve \cite{fpgadts}.

CPro1 fails to replicate FunSearch's 512-size Cap Set solution for $n=8$.  One issue may be a vast difference in scale: while CPro1's run generated only 1000 candidate programs, FunSearch's 512 result was found in 4 of 140 runs, each of which generated 2.5 million candidate programs \cite{funsearch}.  Other attempts to replicate FunSearch's 512 result with LLM-generated code have also failed \cite{capsetrepro1,capsetrepro2}.   

CPro1's solutions are built by randomized heuristics, and usually show little structure.  FunSearch's greedy functions are more likely to yield structure that aids mathematical understanding \cite{funsearch}.

\section{Conclusion}
\label{sec:conclusion}

The protocol CPro1 uses LLMs to generate search heuristics, and using a reasoning LLM it successfully solves open instances of the existence problem for 7 types of combinatorial designs from the \textit{Handbook of Combinatorial Designs} (including open instances for 3 of them that were unsolved by CPro1 with non-reasoning LLMs), 3 more from the Feb. 2025 combinatorial design literature, and also yields state of the art results for the Deletion Codes problem that was previously addressed by FunSearch. The code for CPro1 is available,\footnote{https://github.com/Constructive-Codes/CPro1} and the protocol can be run on additional types of combinatorial designs by supplying a textual definition, a Python verifier, and small collections of parameters for development instances and open instances.  

\clearpage

\bibliographystyle{abbrvnat}
\bibliography{protocol}


\appendix
\clearpage

\section{Appendix: Definitions Used for Prompting}
\label{sec:pseudocode}

\begin{table}
    \centering
  
    \begin{tabular}{| m{0.97\linewidth}| }
    \hline
       A \textbf{"Balanced Incomplete Block Design"} BIBD(v,b,r,k,L) is a pair (V,B) where V is a v-set and B is a collection of b k-subsets of V (blocks) such that each element of V is contained in exactly r blocks and any 2-subset of V is contained in exactly L blocks.\\
The BIBD is represented by a v by b incidence matrix (v rows and b columns) with elements in {0,1}.  The matrix element m\_\{ij\} in the i'th row and j'th column is 1 iff element i is contained in block j.  The sum of each row is r, and the sum of each column is k.  For each pair of distinct elements y and z, sum\_\{j=1\}\textasciicircum \{b\} m\_\{yj\} m\_\{zj\} = L.\\
Given (v,b,r,k,L) we want to find the incidence matrix for a valid BIBD(v,b,r,k,L).
        \\
        \hline
        A \textbf{"Packing Array"} PA(N,k,v) is an N x k array (N rows and k columns), with each entry from the v-set {0,1,...v-1}, so that every N x 2 subarray contains every ordered pair of symbols at most once.  Given (N,k,v), we want to construct PA(N,k,v).\\
        \hline

        An \textbf{"Orthogonal Array"} OA(N,k,s) of size N, degree k, and order s, is a k x N array (k rows and N columns) with entries from the s-set \{0,1,...,s-1\} having the property that in every 2 x N submatrix, every 2 x 1 column vector appears the same number (lambda) of times.  lambda is called the index of the OA, and lambda = N/s\textasciicircum 2.  Given (N,k,s,lambda), we want to construct an OA(N,k,s) with index lambda.\\
        \hline
A "weighing matrix" W(n,w) with parameters (n,w) is an n by n square matrix (n rows and n columns) with entries in \{0,1,-1\} that satisfies W W\textasciicircum T = wI.  That is, W times its transpose is equal to the constant w times the identity matrix I.  The weighing matrix will have w nonzero entries in each row and each column.  And each pair of distinct rows is orthogonal (dot product zero).  Given (n,w), we want to construct "SymmW", a \textbf{symmetric weighing matrix} W(n,w) that satisfies these properties and is also a symmetric matrix.\\
\hline
A "weighing matrix" W(n,w) with parameters (n,w) is an n by n square matrix (n rows and n columns) with entries in \{0,1,-1\} that satisfies W W\textasciicircum T = wI.  That is, W times its transpose is equal to the constant w times the identity matrix I.  The weighing matrix will have w nonzero entries in each row and each column.  And each pair of distinct rows is orthogonal (dot product zero).  Given (n,w), we want to construct "SkewW", a \textbf{skew weighing matrix} W(n,w) that satisfies these properties and is also a skew matrix: that is, W\textasciicircum T = -W.\\
\hline
A \textbf{"Bhaskar Rao Design"} BRD(v,b,r,k,L) is represented by a v by b array (v rows and b columns) with elements a\_\{ij\} in \{-1,0,1\}.  Each row contains exactly r nonzero elements, and each column contains exactly k nonzero elements.  For any pair of distinct rows, the list of pairwise element products must contain -1 L/2 times, and must contain +1 L/2 times.  That is, for distinct rows f and g, the set of pairwise element products \{a\_\{fj\}*a\_\{gj\}\} contains -1 L/2 times, +1 L/2 times, and 0 b-L times.\\
Given (v,b,r,k,L) we want to find a valid BRD(v,b,r,k,L).
\\
\hline
A \textbf{"Balanced Ternary Design"} BTD(V,B;p1,p2,R;K,L) is an arrangement of V elements into B multisets, or blocks, each of cardinality K (K<=V) satisfying:\\
1. Each element appears R=p1 + 2*p2 times altogether, with multiplicity one in exactly p1 blocks and multiplicity two in exactly p2 blocks.\\
2. Every pair of distinct elements appears L times; that is, if m\_\{vb\} is the multiplicity of the v'th element in the b'th block, then for every pair of distinct elements v and w, sum\_\{b=1\}\textasciicircum \{B\} m\_\{vb\} m\_\{wb\} = L.\\
The BTD is represented by a V by B incidence matrix with elements in {0,1,2}.  The matrix element m\_\{vb\} in the v'th row and b'th column is the multiplicity of the v'th element in the b'th block.  The sum of each row is R, and the sum of each column is K.\\
Given (V,B,p1,p2,R,K,L) we want to find BTD(V,B;p1,p2,R;K,L).
\\
\hline
A \textbf{"Costas Array"} of order n is an n by n array of dots and blanks that satisfies:\\
(1) There are n dots and n(n-1) blanks, with exactly one dot in each row and each column.\\
(2) All the segments between pairs of dots differ in length or slope.\\
We will represent the Costas Array by a one-dimensional list "CA" of length n, that contains a permutation of \{0,1,...,n-1\}.  CA[i] identifies the row for the dot that is in column i of the grid.  Condition (1) is automatically satisfied by this representation.  Condition (2) is satisfied if the tuples (j-i,CA[j]-CA[i]) are unique across all j>i with 0<=i,j<n.\\
Given n, we want to construct the one-dimensional list CA that represents a valid Costas Array of order n.
\\
\hline
\end{tabular}
\vspace*{5pt}
    \caption{\textbf{Definitions for use in prompting}: \textit{Handbook} combinatorial design problems part 1.  Note the definitions use ASCII symbols for the sake of prompting. }
    \label{tab:definitions1}
\end{table}

\begin{table}
    \centering
    \begin{tabular}{| m{0.97\linewidth}| }
    \hline
A \textbf{"Covering"} Cov(t,v,k,n) is a pair (X,B), where X is a v-set of elements and B is a collection of k-subsets of X, such that every t-subset of X occurs in at least one block in B.  B has n blocks, and it is required that t<k.  We represent the Covering by an n by v incidence matrix (n rows and v columns) with elements in \{0,1\}; a 1 in row i column j indicates that block i contains the j'th element.  There are k 1's per row.  Given (t,v,k,n) we want to construct a Cov(t,v,k,n) and provide the incidence matrix.
\\
\hline
A \textbf{"Difference Triangle Set"} (n,k)-DTS is a set X=\{X\_1,...,X\_n\} where for 1<=i<=n, X\_i=\{a\_\{i0\}, a\_\{i1\},..., a\_\{ik\}\} with a\_\{ij\} an integer and with 0 = a\_\{i0\} < a\_\{i1\} <a\_\{i2\} < ... < a\_\{ik\}.  The differences a\_\{il\}-a\_\{ij\} for 1<=i<=n, 0<=j!=l<=k are all distinct and nonzero.  The "scope" of a DTS is the max of all a\_\{ij\} in the DTS.\\
Given (n,k) and s, we want to construct an (n,k)-DTS with scope s.  The (n,k)-DTS should be represented by an n by k+1 array (n rows and k+1 columns) of elements a\_\{ij\} with 1<=i<=n and 0<=j<=k.
\\
\hline
Given a k-tuple (x\_0, x\_1, x\_2, ..., x\_\{k-1\}), elements x\_i, x\_\{i+t\} are t-apart in the k-tuple, where i+t is taken modulo k.\\
A \textbf{"Perfect Mendelsohn Design"} with parameters v and k is denoted as a (v,k)-PMD.  It is a set V=\{0,1,...,v-1\} of size v together with a collection B of blocks of ordered k-tuples of distinct elements from V, such that for every i=1,2,3,...,k-1 each ordered pair (x,y) of distinct elements from V is i-apart in exactly one block.  Note since there are v*(v-1) pairs of distinct elements that must be 1-apart in exactly one block, and each block has k pairs that are 1-apart, the design will contain b=v*(v-1)/k blocks.  We will use b to denote the number of blocks. \\ 
Given (v,k,b), we want to construct a (v,k)-PMD with b blocks.  We will represent the PMD with a b by k array (b rows and k columns), with each element of the array chosen from \{0,1,...,v-1\}.
\\
\hline
An \textbf{"equidistant permutation array"} (EPA) with parameters (n,d,m) can be represented as an m by n matrix (m rows and n columns), where each row is the permutation of the numbers 0 to n-1.  Each pair of distinct rows must differ in exactly d positions.  Given (n,d,m), we want to construct an equidistant permutation array (EPA) with these parameters.
\\
\hline
A \textbf{"Florentine Rectangle"} FR(r,n) is an r x n array (r rows and n columns), with each row having a permutation of the set of symbols S=\{0,1,2,...,n-1\}, such that for any two distinct symbols a and b in S and each m in \{1,2,3,...,n-1\} there is at most one row in which b appears in the position which is m steps to the right of a.  A single row will have n-m pairs of symbols a,b with b being m steps to the right of a; so n-1 pairs with b directly to the right of a, n-2 with b 2 steps to the right of a, and only 1 pair with b n-1 steps to the right of a.  Given (r,n) we want to construct a FR(r,n).\\
\hline
A \textbf{"Circular Florentine Rectangle"} CFR(r,n) is an r x n array (r rows and n columns), with each row having a permutation of the set of symbols S=\{0,1,2,...,n-1\}, such that for any two distinct symbols a and b in S and each m in \{1,2,3,...,n-1\} there is at most one row in which b appears in the position which is m steps to the right of a.  "Steps to the right" is taken circularly - so if a is at position i then b is at position (i+m) mod n.  Given (r,n) we want to construct a CFR(r,n).\\
\hline
A \textbf{"Tuscan-2 Square"} T2(n) of size n is an n x n array (n rows and n columns), with each row having a permutation of the set of symbols S=\{0,1,2,...,n-1\}, such that any two distinct symbols a and b in S have exactly one row in which b appears in the position directly to the right of a, and at most one row in which b appears two positions to the right of a (with one symbol between).  Given n, we want to construct a T2(n).
\\
\hline
A \textbf{"Supersimple Balanced Incomplete Block Design"} SBIBD(v,b,r,k,L) is a pair (V,B) where V is a v-set and B is a collection of b k-subsets of V (blocks) such that each element of V is contained in exactly r blocks, any 2-subset of V is contained in exactly L blocks, and any two distinct blocks have at most two elements in common.\\
The SBIBD is represented by a v by b incidence matrix (v rows and b columns) with elements in \{0,1\}.  The matrix element m\_\{ij\} in the i'th row and j'th column is 1 iff element i is contained in block j.  The sum of each row is r, and the sum of each column is k.  For each pair of distinct elements y and z, sum\_\{j=1\}\textasciicircum \{b\} m\_\{yj\} m\_\{zj\} = L.  For each pair of distinct blocks g and h, sum\_\{i=1\}\textasciicircum \{v\} m\_\{ig\} m\_\{ih\} <= 2.\\
Given (v,b,r,k,L) we want to find the incidence matrix for a valid SBIBD(v,b,r,k,L).\\
\hline
\end{tabular}
\vspace*{5pt}
    \caption{\textbf{Definitions for use in prompting}: \textit{Handbook} combinatorial design problems part 2.  Note the definitions use ASCII symbols for the sake of prompting. }
    \label{tab:definitions2}
\end{table}

\begin{table}
    \centering
  
    \begin{tabular}{| m{0.97\linewidth}| }
    \hline
  An "(n,R)-\textbf{Covering Sequence}" (abbreviated "(n,R)-CS") of length L is a cyclic sequence x\_0,x\_1,...,x\_\{L-1\} of length L, over the binary alphabet (x\_i is in \{0,1\} for all j) such that for any length-n binary word y\_0,y\_1,...,y\_\{n-1\} there exists a j such that subsequence x\_j,x\_\{(j+1) mod L\},x\_\{(j+2) mod L\}...,x\_\{(j+n-1) mod L\} of length n is Hamming distance at most R away from the word.  That is, y\_0,...,y\_\{n-1\} and x\_j,...,x\_\{(j+n-1) mod L\} differ in at most R positions.  For our purposes, n <= 16 and R <= 3 and L <= 1200.  Given (n,R,L) we want to construct a (n,R)-CS of length at most L.  Output (n,R)-CS as a list of values in \{0,1\} separated by spaces, all on one line.\\
        \hline
        The Johnson Graph J(N,k) with k<=N/2 is the graph whose vertices are k-element subsets of [N]=\{1,2,...,N\}, with two subsets connected by an edge if their intersection has size exactly k-1.  A \textbf{"Johnson Clique Cover"} JCC(N,k,C) of size C is a set of C cliques in J(N,k), such that the union of these cliques includes all vertices in J(N,k).  Note the cliques in the clique cover do not need to be disjoint; they may share vertices.  For our purposes, N<=15 and C<800.  Given (N,k,C) we want to construct a JCC(N,k,C).  It is a theorem that it suffices to consider clique covers that consist only of maximal cliques, and that all maximal cliques of J(N,k) are either type 0 or type 1, defined as follows.  A type 0 clique specifies a k-1 element subset S of [N], and consists of all vertices corresponding to the subset S plus x, for each x that is an element of [N] that is not in S.  A type 1 clique specifies k+1 elements in [N], and consists of all vertices corresponding to the subset S excluding x, for each x that is an element of S.  Specify a clique as a space-separated list of integers where the first integer is the type (0 or 1) and the remaining integers specify the elements of S.  For example in J(5,2) the type 0 clique "0 1" consists of vertices for the subsets {1,2}, {1,3}, {1,4}, and {1,5}; and the type 1 clique "1 3 4 5" consists of vertices for the subsets \{4,5\}, \{3,4\}, and \{3,5\}.  Output the clique cover as C lines, with one clique per line, where each line is a space-separated list of integers starting with the type (0 or 1).\\
        \hline
A "Steiner Quadruple System" (or SQS) of order v consists of a set of blocks, with each block containing 4 elements of the set V=\{0,1,...,v-1\}, such that each subset of 3 elements of V is contained in exactly one block.  There are v*(v-1)*(v-2)/6 subsets of 3 elements of V, and each block covers 4 of these subsets, so the SQS will have v*(v-1)*(v-2)/24 blocks.  A \textbf{"Uniform Nested Steiner Quadruple System"} of order v splits each block into two "ND-pairs" of two elements each, such that each distinct ND-pair appears the same number of times.  Set p denote the number of distinct ND-pairs that appear among the blocks; it may be that each of the v*(v-1)/2 subsets of 2 elements from V appears as an ND-pair so that p = v*(v-1)/2, or it may be that some subsets of 2 elements from V don't appear as an ND-pair and then we have p < v*(v-1)/2.  There are v*(v-1)*(v-2)/12 ND-pairs, so each of the p distinct ND-pairs appears v*(v-1)*(v-2)/(12*p) times.  We call such a design a UNSQS(v,p).  Given (v,p) we want to construct a UNSQS(v,p).  For our purposes, 8<=v<=58 and 28<=p<=1260.  Output the UNSQS(v,p) as v*(v-1)*(v-2)/24 lines, one block per line, with each line having a space-separated list of 4 numbers identifying the elements of V in the block.  The first ND-pair in a block should be the first two elements listed for the block, and the second ND-pair is the last two elements listed.  Order does not matter within an ND-pair.\\
\hline
A "\textbf{Covering Array} of Strength 3" CA3(N,k,v) is an N x k array (N rows and k columns), with each entry from the v-set of symbols \{0,1,...v-1\}, so that every N x 3 subarray contains every ordered triple of symbols at least once.  Given (N,k,v), we want to construct CA3(N,k,v).  For our purposes, N<1000, k<1000, and v<6.  Output the CA3(N,k,v) as N lines with one row per line, with each line a space-separated list of k columns where each column is an integer in \{0,1,...,v-1\}.
\\
\hline
\end{tabular}
\vspace*{5pt}
    \caption{\textbf{Definitions for use in prompting}: Feb. 2025 combinatorial design problems.  Note the definitions use ASCII symbols for the sake of prompting. }
    \label{tab:definitions3}
\end{table}

\begin{table}
    \centering
  
    \begin{tabular}{| m{0.97\linewidth}| }
    \hline
  A \textbf{"Cap Set"} CS(n,s) is a subset S of Z\_3\textasciicircum \{n\}, such that S has at least s distinct points, and no three points \{x,y,z\} in S satisfy x+y+z=0 (vector addition over Z\_3\textasciicircum \{n\}, so addition is taken modulo 3).  We represent the cap set by an array with at least s rows, and n columns in each row.  Elements of the array are from Z\_3=\{0,1,2\}, and each row represents a point in Z\_3\textasciicircum \{n\} which is an element of S.  Given (n,s) we want to construct a Cap Set CS(n,s).\\
        \hline
        A \textbf{"Deletion Code"} DC(n,s,m) with parameters (n,s,m) is a set m binary words of length n, such that any two distinct words from the set do not share any any length n-s word obtained by deleting s bits from each of the two words.  Given a length n word x, let D(x) be the set of length n-s words obtained from x by deleting two distinct bits (at positions which need not be adjacent).  Note D(x) has (n choose s) members.  Our requirement is that, for any two distinct words x and y in DC(n,s,m), D(x) and D(y) have the empty intersection.  Given (n,s,m) we want to construct a DC(n,s,m).  For our purposes, 7<=n<=16, s is 2 or 3, and m<250.  Output the DC(n,s,m) as an m by n array, where each row is a space-separated list of n bits in \{0,1\} representing one word in the Deletion Code.\\
        \hline
\end{tabular}
\vspace*{5pt}
    \caption{\textbf{Definitions for use in prompting}: FunSearch problems.  Note the definitions use ASCII symbols for the sake of prompting. }
    \label{tab:definitions4}
\end{table}

\clearpage
\section{Appendix: Solutions to Open Instances of Combinatorial Design Problems}
\label{sec:solutions}

Figures~\ref{fig:arrays1}~to~\ref{fig:arrays7} show verified designs that resolve open instances of combinatorial design problems as noted in Table~\ref{tab:mainresults}.  Each of these was constructed by code that was generated by protocol CPro1, and that code is available in the github repository \footnote{https://github.com/Constructive-Codes/CPro1}.  One example is provided here for each type of combinatorial design -- remaining solutions are in the repository.

\begin{figure*}[h!]
\centering
\begingroup              
    \setlength{\arraycolsep}{3pt}%
\[
\fbox{
$\begin{array}{*{22}{r}}
0 & 0 & 0 & -1 & 0 & 1 & -1 & 0 & -1 & 1 & -1 & -1 & -1 & 1 & 1 & -1 & 0 & 1 & 1 & -1 & 1 & 1\\ 
0 & 0 & 0 & 1 & 0 & -1 & 0 & 1 & 1 & 1 & 1 & -1 & 0 & -1 & -1 & -1 & -1 & 1 & -1 & -1 & 1 & 1\\ 
0 & 0 & -1 & -1 & 0 & 0 & 0 & 1 & -1 & -1 & 1 & 1 & -1 & 1 & -1 & -1 & 0 & -1 & -1 & -1 & -1 & 1\\ 
-1 & 1 & -1 & 0 & -1 & 0 & -1 & 0 & -1 & -1 & -1 & -1 & -1 & -1 & -1 & 1 & 0 & 1 & -1 & 1 & 0 & 0\\ 
0 & 0 & 0 & -1 & -1 & -1 & 1 & 0 & -1 & -1 & 1 & 1 & 0 & -1 & 1 & 1 & -1 & 1 & 1 & -1 & 1 & 0\\ 
1 & -1 & 0 & 0 & -1 & -1 & -1 & 0 & -1 & 1 & -1 & 1 & 0 & -1 & -1 & -1 & -1 & -1 & 1 & 1 & 0 & 0\\ 
-1 & 0 & 0 & -1 & 1 & -1 & -1 & 0 & 1 & 1 & 1 & 1 & -1 & 1 & -1 & 1 & 0 & 1 & 1 & 1 & 0 & 0\\ 
0 & 1 & 1 & 0 & 0 & 0 & 0 & -1 & 1 & -1 & -1 & 1 & 0 & -1 & -1 & -1 & 1 & 1 & 1 & -1 & -1 & 1\\ 
-1 & 1 & -1 & -1 & -1 & -1 & 1 & 1 & 0 & 1 & 0 & -1 & 1 & 0 & 0 & -1 & 1 & 0 & 1 & 0 & -1 & -1\\ 
1 & 1 & -1 & -1 & -1 & 1 & 1 & -1 & 1 & 0 & 0 & 0 & 1 & 1 & -1 & 0 & -1 & 0 & 0 & 1 & 1 & 1\\ 
-1 & 1 & 1 & -1 & 1 & -1 & 1 & -1 & 0 & 0 & -1 & 0 & -1 & 0 & 0 & -1 & -1 & -1 & -1 & 0 & 1 & -1\\ 
-1 & -1 & 1 & -1 & 1 & 1 & 1 & 1 & -1 & 0 & 0 & 0 & 1 & -1 & -1 & 0 & 1 & 0 & 0 & 1 & 1 & 1\\ 
-1 & 0 & -1 & -1 & 0 & 0 & -1 & 0 & 1 & 1 & -1 & 1 & 1 & -1 & 1 & 1 & 0 & -1 & -1 & -1 & 0 & 1\\ 
1 & -1 & 1 & -1 & -1 & -1 & 1 & -1 & 0 & 1 & 0 & -1 & -1 & 0 & 0 & 1 & 1 & 0 & -1 & 0 & -1 & 1\\ 
1 & -1 & -1 & -1 & 1 & -1 & -1 & -1 & 0 & -1 & 0 & -1 & 1 & 0 & -1 & 0 & 1 & 0 & 0 & -1 & 1 & -1\\ 
-1 & -1 & -1 & 1 & 1 & -1 & 1 & -1 & -1 & 0 & -1 & 0 & 1 & 1 & 0 & 0 & -1 & 1 & 0 & 0 & -1 & 1\\ 
0 & -1 & 0 & 0 & -1 & -1 & 0 & 1 & 1 & -1 & -1 & 1 & 0 & 1 & 1 & -1 & 1 & 1 & -1 & 1 & 1 & 0\\ 
1 & 1 & -1 & 1 & 1 & -1 & 1 & 1 & 0 & 0 & -1 & 0 & -1 & 0 & 0 & 1 & 1 & -1 & 1 & 0 & 1 & 1\\ 
1 & -1 & -1 & -1 & 1 & 1 & 1 & 1 & 1 & 0 & -1 & 0 & -1 & -1 & 0 & 0 & -1 & 1 & 0 & 0 & -1 & -1\\ 
-1 & -1 & -1 & 1 & -1 & 1 & 1 & -1 & 0 & 1 & 0 & 1 & -1 & 0 & -1 & 0 & 1 & 0 & 0 & -1 & 1 & -1\\ 
1 & 1 & -1 & 0 & 1 & 0 & 0 & -1 & -1 & 1 & 1 & 1 & 0 & -1 & 1 & -1 & 1 & 1 & -1 & 1 & 0 & 0\\ 
1 & 1 & 1 & 0 & 0 & 0 & 0 & 1 & -1 & 1 & -1 & 1 & 1 & 1 & -1 & 1 & 0 & 1 & -1 & -1 & 0 & -1
\end{array}$}
\]
\endgroup
\caption{Symmetric Weighing Matrix with n=22 w=16}
\label{fig:arrays1}
\end{figure*}

\begin{figure*}[h!]
\centering
\[
\fbox{
$\begin{array}{*{15}{r}}
1 & 0 & 0 & 1 & 0 & -1 & 1 & 0 & 0 & 0 & 0 & 0 & 1 & 0 & 0\\ 
1 & 0 & 0 & 0 & 1 & 0 & 0 & 1 & 0 & 0 & 0 & 0 & -1 & 0 & -1\\ 
0 & 0 & 0 & 0 & 0 & 0 & 0 & 1 & -1 & 1 & -1 & 0 & 1 & 0 & 0\\ 
0 & 0 & -1 & 0 & 0 & 0 & 0 & 0 & 1 & 0 & 1 & 0 & 0 & -1 & 1\\ 
0 & 0 & 0 & 0 & 1 & 1 & 0 & 0 & 0 & 1 & 0 & 0 & 1 & -1 & 0\\ 
0 & -1 & 0 & -1 & 0 & -1 & 0 & 1 & 0 & 0 & 0 & 0 & 0 & 1 & 0\\ 
0 & -1 & 0 & 0 & -1 & 0 & 0 & 0 & 0 & 0 & 1 & 0 & 1 & 0 & -1\\ 
0 & 1 & -1 & 1 & 0 & 0 & 0 & 0 & 0 & 0 & 1 & 0 & 1 & 0 & 0\\ 
0 & 0 & -1 & -1 & 0 & 1 & 0 & 0 & 0 & 0 & 0 & 1 & 0 & 0 & -1\\ 
0 & 0 & 0 & 0 & 0 & 0 & 1 & 1 & 0 & 0 & 1 & 1 & 0 & 1 & 0\\ 
0 & 0 & 0 & 1 & -1 & 0 & 0 & 0 & 1 & 1 & -1 & 0 & 0 & 0 & 0\\ 
0 & 1 & 0 & 0 & 0 & -1 & -1 & -1 & 0 & 0 & 0 & 0 & 0 & 0 & -1\\ 
1 & 0 & 0 & 0 & 0 & 0 & 1 & 0 & 1 & 0 & 0 & 0 & -1 & 1 & 0\\ 
0 & 0 & -1 & 0 & 0 & 0 & 0 & 0 & 0 & 0 & 0 & -1 & -1 & -1 & -1\\ 
1 & -1 & 0 & 1 & 0 & 0 & 0 & 0 & 0 & -1 & 0 & 0 & 0 & -1 & 0\\ 
1 & 1 & 0 & 0 & 1 & 0 & 0 & 0 & 0 & -1 & 0 & 1 & 0 & 0 & 0\\ 
0 & 0 & 0 & 0 & 0 & 1 & 0 & 0 & 1 & -1 & 0 & -1 & 0 & 0 & 1\\ 
0 & -1 & -1 & 0 & 0 & -1 & 0 & 0 & 1 & 1 & 0 & 0 & 0 & 0 & 0\\ 
1 & 0 & 0 & 0 & 0 & 1 & 0 & 0 & 1 & 0 & -1 & 1 & 0 & 0 & 0\\ 
-1 & 0 & 1 & 0 & 1 & 0 & 0 & 0 & 1 & 0 & 0 & 0 & 0 & -1 & 0\\ 
1 & 0 & 0 & 0 & 0 & 0 & -1 & 0 & 0 & 1 & 1 & 0 & 0 & 0 & 1\\ 
0 & 0 & 0 & 1 & 0 & 0 & 1 & 0 & -1 & 0 & 0 & -1 & -1 & 0 & 0\\ 
1 & 0 & 0 & 0 & -1 & 1 & -1 & 0 & 0 & 0 & 0 & 0 & 0 & 0 & -1\\ 
0 & 1 & 0 & -1 & -1 & 0 & 0 & 0 & -1 & 0 & 0 & 0 & 0 & 0 & 1\\ 
0 & 0 & 1 & 0 & 0 & 0 & 1 & -1 & 0 & -1 & 0 & 0 & 1 & 0 & 0\\ 
0 & 0 & 1 & 0 & -1 & -1 & 0 & 0 & 0 & 0 & 0 & 0 & -1 & -1 & 0\\ 
1 & 0 & 1 & -1 & 0 & 0 & 0 & -1 & 0 & 1 & 0 & 0 & 0 & 0 & 0\\ 
0 & 0 & 0 & 1 & 0 & 0 & 0 & 0 & 0 & 1 & 0 & 1 & -1 & 0 & 1\\ 
0 & 0 & 0 & 0 & 1 & -1 & 0 & -1 & 0 & 0 & -1 & 1 & 0 & 0 & 0\\ 
0 & 0 & 0 & -1 & 0 & 0 & 0 & 1 & 0 & 0 & -1 & 0 & 0 & -1 & 1\\ 
0 & 0 & -1 & 0 & 0 & -1 & -1 & 0 & 0 & -1 & -1 & 0 & 0 & 0 & 0\\ 
-1 & -1 & -1 & 0 & 0 & 0 & 0 & -1 & 0 & 0 & 0 & 1 & 0 & 0 & 0\\ 
0 & -1 & 0 & 0 & 0 & 1 & 0 & -1 & -1 & 0 & 0 & 0 & -1 & 0 & 0\\ 
-1 & 0 & 1 & 0 & 0 & 0 & 0 & 1 & 1 & 0 & 0 & 0 & 0 & 0 & -1\\ 
0 & 0 & 0 & 1 & 1 & 0 & -1 & 1 & -1 & 0 & 0 & 0 & 0 & 0 & 0\\ 
0 & 0 & -1 & -1 & 1 & 0 & 1 & 0 & 0 & 0 & 0 & -1 & 0 & 0 & 0\\ 
0 & 1 & -1 & 0 & -1 & 0 & 1 & 0 & 0 & 0 & -1 & 0 & 0 & 0 & 0\\ 
0 & 1 & 0 & 0 & 0 & 0 & 1 & 0 & 0 & 1 & 0 & 0 & 0 & -1 & -1\\ 
1 & 0 & 0 & -1 & 0 & -1 & 0 & 0 & 0 & 0 & 1 & 0 & 0 & -1 & 0\\ 
0 & -1 & 0 & 0 & 0 & 0 & 1 & 0 & -1 & 0 & 0 & 1 & 0 & -1 & 0\\ 
1 & -1 & 0 & 0 & 0 & 0 & 0 & 0 & 0 & 0 & -1 & -1 & 1 & 0 & 0\\ 
0 & 0 & 0 & 0 & -1 & 0 & 0 & 1 & 0 & -1 & 0 & 1 & 0 & -1 & 0
\end{array}$}
\]
\caption{Bhaskar Rao Design with parameters (15,42,14,5,4).  Note the transpose is shown.}
\label{fig:arrays2}
\end{figure*}

\begin{figure*}[h!]
\centering
\[
\fbox{
$\begin{array}{*{22}{r}}
0 & 0 & 2 & 0 & 0 & 1 & 0 & 1 & 0 & 1 & 1 & 0 & 1 & 1 & 0 & 0 & 1 & 0 & 0 & 0 & 1 & 1\\ 
2 & 1 & 1 & 1 & 0 & 1 & 0 & 0 & 0 & 0 & 0 & 1 & 1 & 0 & 0 & 1 & 0 & 0 & 1 & 0 & 0 & 1\\ 
1 & 0 & 1 & 0 & 0 & 1 & 2 & 1 & 1 & 0 & 1 & 0 & 0 & 0 & 0 & 1 & 0 & 1 & 0 & 1 & 0 & 0\\ 
1 & 0 & 0 & 0 & 1 & 0 & 0 & 0 & 1 & 0 & 2 & 1 & 1 & 1 & 1 & 0 & 0 & 0 & 0 & 1 & 0 & 1\\ 
0 & 1 & 0 & 0 & 1 & 2 & 1 & 0 & 0 & 1 & 0 & 1 & 1 & 1 & 1 & 0 & 0 & 1 & 0 & 0 & 0 & 0\\ 
0 & 1 & 0 & 0 & 0 & 1 & 0 & 1 & 1 & 0 & 0 & 1 & 1 & 0 & 0 & 0 & 1 & 0 & 1 & 2 & 1 & 0\\ 
1 & 0 & 0 & 0 & 1 & 0 & 1 & 2 & 0 & 1 & 0 & 1 & 0 & 0 & 1 & 0 & 1 & 0 & 1 & 0 & 0 & 1\\ 
0 & 0 & 1 & 1 & 0 & 0 & 1 & 0 & 0 & 0 & 0 & 1 & 0 & 2 & 1 & 1 & 1 & 0 & 1 & 1 & 0 & 0\\ 
1 & 1 & 1 & 0 & 1 & 0 & 0 & 0 & 2 & 1 & 0 & 0 & 0 & 1 & 0 & 0 & 1 & 1 & 1 & 0 & 0 & 0\\ 
0 & 0 & 1 & 1 & 2 & 0 & 1 & 0 & 0 & 0 & 0 & 0 & 1 & 0 & 0 & 0 & 0 & 1 & 1 & 1 & 1 & 1\\ 
1 & 1 & 0 & 1 & 0 & 0 & 1 & 0 & 0 & 0 & 1 & 0 & 1 & 0 & 1 & 0 & 2 & 1 & 0 & 0 & 1 & 0\\ 
0 & 0 & 0 & 1 & 0 & 1 & 0 & 0 & 1 & 1 & 0 & 0 & 0 & 0 & 1 & 1 & 1 & 1 & 0 & 1 & 0 & 2\\ 
0 & 0 & 0 & 1 & 0 & 0 & 1 & 0 & 1 & 2 & 1 & 1 & 1 & 0 & 0 & 1 & 0 & 0 & 1 & 0 & 1 & 0\\ 
0 & 1 & 0 & 0 & 0 & 0 & 0 & 1 & 0 & 0 & 1 & 1 & 0 & 1 & 0 & 1 & 0 & 2 & 1 & 0 & 1 & 1\\ 
0 & 2 & 1 & 1 & 1 & 0 & 0 & 1 & 0 & 1 & 1 & 0 & 0 & 0 & 1 & 1 & 0 & 0 & 0 & 1 & 0 & 0\\ 
1 & 0 & 0 & 1 & 1 & 1 & 0 & 1 & 1 & 0 & 0 & 0 & 0 & 1 & 1 & 1 & 0 & 0 & 0 & 0 & 2 & 0
\end{array}$}
\]
\caption{Balanced Ternary Design with parameters (16,22;9,1,11;8,5)}
\label{fig:arrays3}
\end{figure*}

\begin{figure*}[h!]
\centering
\begingroup              
    \setlength{\arraycolsep}{0.2pt}%
\[
\fbox{
$\begin{array}{*{71}{r}}
0 & 1 & 0 & 1 & 1 & 1 & 0 & 1 & 0 & 1 & 0 & 0 & 1 & 1 & 1 & 0 & 0 & 0 & 0 & 1 & 0 & 0 & 0 & 0 & 0 & 0 & 1 & 1 & 0 & 1 & 1 & 0 & 0 & 1 & 1 & 0 & 0 & 1 & 1 & 0 & 1 & 1 & 0 & 1 & 0 & 0 & 0 & 1 & 0 & 1 & 0 & 1 & 1 & 0 & 0 & 0 & 1 & 1 & 1 & 1 & 0 & 1 & 1 & 1 & 1 & 1 & 1 & 0 & 0 & 1 & 0
\end{array}$}
\]
\endgroup
\caption{Covering Sequence with n=9 R=1 L=71}
\label{fig:arrays4}
\end{figure*}

\begin{figure*}[h!]
\centering
  \begin{minipage}[t]{0.47\linewidth}
    \begin{topframe}
\[
\begin{array}{*{6}{r}}
0 & 2 & 3 & 4\\ 
1 & 1 & 3 & 4 & 9 & 13\\ 
0 & 3 & 9 & 12\\ 
0 & 1 & 2 & 3\\ 
1 & 1 & 3 & 5 & 6 & 11\\ 
0 & 5 & 7 & 10\\ 
0 & 2 & 8 & 9\\ 
0 & 2 & 5 & 12\\ 
0 & 6 & 7 & 8\\ 
0 & 3 & 6 & 13\\ 
0 & 2 & 6 & 9\\ 
0 & 2 & 9 & 13\\ 
0 & 1 & 11 & 13\\ 
0 & 1 & 6 & 8\\ 
0 & 9 & 10 & 11\\ 
0 & 1 & 4 & 6\\ 
1 & 5 & 9 & 10 & 12 & 13\\ 
0 & 3 & 6 & 10\\ 
1 & 1 & 2 & 6 & 10 & 11\\ 
0 & 3 & 10 & 13\\ 
1 & 1 & 6 & 7 & 9 & 13\\ 
1 & 6 & 8 & 9 & 12 & 13\\ 
1 & 4 & 5 & 6 & 11 & 12\\ 
1 & 3 & 7 & 9 & 11 & 13\\ 
0 & 6 & 7 & 11\\ 
0 & 2 & 8 & 12\\ 
0 & 1 & 2 & 7\\ 
1 & 3 & 5 & 7 & 11 & 12\\ 
0 & 4 & 5 & 9\\ 
0 & 4 & 7 & 12\\ 
0 & 3 & 8 & 13\\ 
0 & 6 & 8 & 11\\ 
0 & 5 & 7 & 9\\ 
0 & 2 & 8 & 10\\ 
0 & 6 & 9 & 12\\ 
0 & 1 & 4 & 10\\ 
0 & 6 & 10 & 13\\ 
1 & 2 & 5 & 7 & 11 & 13\\ 
0 & 4 & 7 & 9\\ 
0 & 2 & 5 & 6\\ 
1 & 1 & 5 & 6 & 7 & 12\\ 
0 & 11 & 12 & 13\\ 
0 & 7 & 12 & 13\\ 
1 & 1 & 3 & 8 & 10 & 11\\ 
1 & 1 & 2 & 4 & 9 & 12\\ 
0 & 2 & 6 & 13\\ 
0 & 7 & 8 & 11\\ 
0 & 8 & 10 & 13\\ 
0 & 4 & 7 & 13\\ 
1 & 2 & 7 & 9 & 11 & 12\\ 
0 & 1 & 3 & 7\\ 
0 & 3 & 5 & 8\\ 
0 & 2 & 10 & 12\\ 
\end{array}
\]
\end{topframe}
\end{minipage}
\hfill
\begin{minipage}[t]{0.47\linewidth}
    \begin{botframe}
      \[
      \begin{array}{*{6}{r}}
0 & 1 & 9 & 11\\ 
0 & 3 & 8 & 9\\ 
0 & 8 & 10 & 12\\ 
0 & 1 & 7 & 10\\ 
0 & 1 & 5 & 9\\ 
1 & 2 & 3 & 5 & 9 & 10\\ 
0 & 3 & 5 & 13\\ 
0 & 3 & 10 & 12\\ 
0 & 2 & 10 & 13\\ 
0 & 1 & 12 & 13\\ 
0 & 7 & 10 & 11\\ 
0 & 4 & 12 & 13\\ 
1 & 2 & 3 & 6 & 8 & 9\\ 
1 & 5 & 6 & 8 & 9 & 10\\ 
1 & 4 & 5 & 8 & 11 & 13\\ 
0 & 1 & 5 & 10\\ 
0 & 2 & 6 & 12\\ 
0 & 2 & 4 & 11\\ 
0 & 4 & 9 & 10\\ 
1 & 3 & 4 & 7 & 8 & 10\\ 
1 & 2 & 4 & 5 & 7 & 8\\ 
0 & 3 & 6 & 9\\ 
1 & 3 & 4 & 5 & 6 & 7\\ 
0 & 7 & 9 & 10\\ 
0 & 2 & 8 & 13\\ 
1 & 1 & 7 & 8 & 9 & 12\\ 
1 & 4 & 8 & 9 & 11 & 12\\ 
0 & 6 & 10 & 12\\ 
1 & 1 & 5 & 7 & 8 & 13\\ 
0 & 2 & 3 & 7\\ 
0 & 8 & 9 & 13\\ 
1 & 1 & 3 & 4 & 5 & 12\\ 
0 & 3 & 8 & 12\\ 
0 & 5 & 9 & 11\\ 
0 & 1 & 11 & 12\\ 
0 & 1 & 4 & 8\\ 
0 & 1 & 9 & 10\\ 
1 & 4 & 6 & 9 & 11 & 13\\ 
0 & 3 & 4 & 11\\ 
0 & 2 & 3 & 11\\ 
0 & 4 & 6 & 8\\ 
1 & 1 & 2 & 5 & 8 & 11\\ 
1 & 2 & 3 & 9 & 12 & 13\\ 
0 & 3 & 6 & 12\\ 
0 & 4 & 10 & 11\\ 
0 & 5 & 10 & 11\\ 
0 & 4 & 5 & 10\\ 
1 & 1 & 4 & 5 & 7 & 11\\ 
0 & 5 & 8 & 12\\ 
1 & 1 & 2 & 4 & 5 & 13\\ 
1 & 2 & 4 & 6 & 7 & 10\\ 
0 & 5 & 6 & 13
\end{array}
\]
\end{botframe}
\end{minipage}
\caption{Johnson Clique Cover with N=13 k=4 C=105}
\label{fig:arrays5}
\end{figure*}

\begin{figure*}[h!]
\centering
  \begin{minipage}[t]{0.47\linewidth}
    \begin{topframe}
\[
\begin{array}{*{4}{r}}
0 & 1 & 2 & 3\\ 
0 & 4 & 1 & 5\\ 
0 & 6 & 1 & 7\\ 
0 & 9 & 1 & 8\\ 
0 & 10 & 1 & 11\\ 
0 & 1 & 12 & 13\\ 
0 & 4 & 2 & 6\\ 
0 & 7 & 2 & 5\\ 
0 & 10 & 2 & 8\\ 
0 & 2 & 9 & 12\\ 
0 & 2 & 11 & 13\\ 
0 & 8 & 3 & 4\\ 
0 & 5 & 3 & 9\\ 
0 & 3 & 6 & 10\\ 
0 & 3 & 7 & 13\\ 
0 & 11 & 3 & 12\\ 
0 & 11 & 4 & 7\\ 
0 & 13 & 4 & 9\\ 
0 & 12 & 4 & 10\\ 
0 & 5 & 6 & 12\\ 
0 & 8 & 5 & 11\\ 
0 & 13 & 5 & 10\\ 
0 & 6 & 8 & 13\\ 
0 & 9 & 6 & 11\\ 
0 & 12 & 7 & 8\\ 
0 & 7 & 9 & 10\\ 
1 & 4 & 2 & 7\\ 
1 & 2 & 5 & 6\\ 
1 & 2 & 8 & 11\\ 
1 & 13 & 2 & 9\\ 
1 & 10 & 2 & 12\\ 
1 & 4 & 3 & 10\\ 
1 & 3 & 5 & 12\\ 
1 & 9 & 3 & 6\\ 
1 & 3 & 7 & 11\\ 
1 & 8 & 3 & 13\\ 
1 & 6 & 4 & 13\\ 
1 & 12 & 4 & 8\\ 
1 & 9 & 4 & 11\\ 
1 & 7 & 5 & 8\\ 
1 & 5 & 9 & 10\\ 
1 & 11 & 5 & 13\\ 
1 & 10 & 6 & 8\\ 
1 & 6 & 11 & 12\\ 
1 & 12 & 7 & 9\\ 
1 & 13 & 7 & 10\\ 
\end{array}
\]
\end{topframe}
\end{minipage}
\hfill
\begin{minipage}[t]{0.47\linewidth}
    \begin{botframe}
      \[
      \begin{array}{*{4}{r}}
2 & 13 & 3 & 4\\ 
2 & 8 & 3 & 5\\ 
2 & 11 & 3 & 6\\ 
2 & 3 & 7 & 12\\ 
2 & 10 & 3 & 9\\ 
2 & 4 & 5 & 10\\ 
2 & 4 & 8 & 9\\ 
2 & 11 & 4 & 12\\ 
2 & 5 & 9 & 11\\ 
2 & 12 & 5 & 13\\ 
2 & 9 & 6 & 7\\ 
2 & 6 & 8 & 12\\ 
2 & 10 & 6 & 13\\ 
2 & 13 & 7 & 8\\ 
2 & 7 & 10 & 11\\ 
3 & 11 & 4 & 5\\ 
3 & 12 & 4 & 6\\ 
3 & 7 & 4 & 9\\ 
3 & 13 & 5 & 6\\ 
3 & 5 & 7 & 10\\ 
3 & 7 & 6 & 8\\ 
3 & 8 & 9 & 12\\ 
3 & 8 & 10 & 11\\ 
3 & 11 & 9 & 13\\ 
3 & 10 & 12 & 13\\ 
4 & 6 & 5 & 7\\ 
4 & 5 & 8 & 13\\ 
4 & 12 & 5 & 9\\ 
4 & 8 & 6 & 11\\ 
4 & 10 & 6 & 9\\ 
4 & 7 & 8 & 10\\ 
4 & 13 & 7 & 12\\ 
4 & 11 & 10 & 13\\ 
5 & 8 & 6 & 9\\ 
5 & 11 & 6 & 10\\ 
5 & 9 & 7 & 13\\ 
5 & 7 & 11 & 12\\ 
5 & 12 & 8 & 10\\ 
6 & 7 & 10 & 12\\ 
6 & 13 & 7 & 11\\ 
6 & 12 & 9 & 13\\ 
7 & 9 & 8 & 11\\ 
8 & 9 & 10 & 13\\ 
8 & 12 & 11 & 13\\ 
9 & 11 & 10 & 12
\end{array}
\]
\end{botframe}
\end{minipage}
\caption{Uniform Nested Steiner Quadruple System with v=14 p=91}
\label{fig:arrays6}
\end{figure*}

\begin{figure*}[h!]
\centering
\[
\fbox{
$\begin{array}{*{12}{r}}
0 & 1 & 1 & 1 & 1 & 1 & 1 & 0 & 0 & 0 & 0 & 1\\ 
1 & 1 & 0 & 0 & 0 & 0 & 0 & 0 & 1 & 1 & 1 & 1\\ 
0 & 1 & 1 & 0 & 1 & 1 & 1 & 1 & 1 & 1 & 0 & 1\\ 
1 & 1 & 1 & 0 & 1 & 1 & 0 & 0 & 0 & 1 & 0 & 0\\ 
1 & 1 & 1 & 1 & 1 & 1 & 1 & 1 & 1 & 1 & 1 & 1\\ 
1 & 1 & 1 & 1 & 0 & 1 & 0 & 1 & 0 & 1 & 0 & 1\\ 
0 & 0 & 0 & 0 & 0 & 1 & 1 & 1 & 1 & 1 & 1 & 1\\ 
0 & 0 & 0 & 1 & 1 & 1 & 1 & 0 & 0 & 0 & 0 & 0\\ 
1 & 1 & 0 & 0 & 0 & 1 & 0 & 0 & 0 & 0 & 0 & 0\\ 
1 & 0 & 0 & 1 & 1 & 1 & 0 & 1 & 1 & 1 & 1 & 1\\ 
1 & 1 & 1 & 1 & 1 & 1 & 1 & 1 & 1 & 0 & 0 & 0\\ 
0 & 0 & 1 & 1 & 0 & 0 & 0 & 0 & 0 & 0 & 1 & 0\\ 
1 & 1 & 1 & 0 & 1 & 1 & 0 & 0 & 1 & 1 & 1 & 1\\ 
1 & 1 & 1 & 1 & 0 & 0 & 0 & 0 & 0 & 0 & 0 & 1\\ 
0 & 0 & 0 & 0 & 0 & 0 & 0 & 0 & 0 & 0 & 0 & 0\\ 
0 & 0 & 0 & 1 & 1 & 0 & 0 & 1 & 1 & 1 & 1 & 0\\ 
1 & 1 & 1 & 1 & 0 & 0 & 0 & 0 & 1 & 1 & 1 & 0\\ 
1 & 0 & 0 & 1 & 1 & 0 & 0 & 1 & 1 & 0 & 0 & 0\\ 
0 & 1 & 0 & 0 & 1 & 1 & 1 & 0 & 1 & 0 & 1 & 0\\ 
1 & 0 & 0 & 0 & 0 & 1 & 0 & 0 & 0 & 1 & 1 & 0\\ 
1 & 0 & 0 & 0 & 0 & 1 & 1 & 1 & 1 & 1 & 0 & 0\\ 
0 & 0 & 0 & 1 & 0 & 1 & 1 & 1 & 0 & 0 & 1 & 1\\ 
0 & 1 & 1 & 1 & 0 & 0 & 1 & 0 & 0 & 0 & 1 & 1\\ 
0 & 0 & 0 & 0 & 0 & 0 & 1 & 1 & 1 & 0 & 0 & 0\\ 
0 & 0 & 1 & 1 & 1 & 1 & 0 & 1 & 1 & 1 & 0 & 0\\ 
0 & 0 & 0 & 0 & 0 & 0 & 0 & 0 & 0 & 1 & 1 & 1\\ 
1 & 0 & 1 & 0 & 1 & 0 & 0 & 0 & 0 & 1 & 0 & 1\\ 
1 & 1 & 1 & 0 & 0 & 1 & 1 & 1 & 1 & 0 & 0 & 1\\ 
0 & 0 & 1 & 0 & 1 & 0 & 1 & 0 & 0 & 1 & 0 & 0\\ 
1 & 1 & 0 & 0 & 0 & 1 & 1 & 0 & 1 & 0 & 1 & 1\\ 
0 & 0 & 1 & 1 & 1 & 0 & 0 & 0 & 1 & 1 & 0 & 1\\ 
0 & 1 & 0 & 0 & 0 & 1 & 1 & 0 & 0 & 0 & 0 & 1\\ 
0 & 1 & 1 & 0 & 0 & 0 & 1 & 0 & 1 & 1 & 0 & 0\\ 
0 & 1 & 0 & 1 & 0 & 1 & 0 & 1 & 0 & 1 & 1 & 1\\ 
0 & 0 & 0 & 0 & 0 & 1 & 0 & 1 & 0 & 1 & 0 & 1\\ 
1 & 0 & 1 & 1 & 1 & 1 & 1 & 0 & 0 & 1 & 1 & 0
\end{array}$}
\]
\caption{Deletion Code with n=12 s=2 m=36}
\label{fig:arrays7}
\end{figure*}

\clearpage

\end{document}